\documentclass[conference]{IEEEtran}
\IEEEoverridecommandlockouts

\usepackage{cite}
\usepackage{amsmath,amssymb,amsfonts}

\usepackage{algorithmic}
\usepackage{algorithm}

\usepackage{graphicx}
\usepackage{textcomp}
\usepackage{soul}
\usepackage{xcolor}
\usepackage[utf8]{inputenc}
\usepackage{mathtools}
\usepackage{amsthm}

\usepackage{etoolbox}
\makeatletter
\patchcmd{\@makecaption}
  {\scshape}
  {}
  {}
  {}
\makeatletter
\patchcmd{\@makecaption}
  {\\}
  {.\ }
  {}
  {}
\makeatother
\def\tablename{Table}
\begin{document}

\title{Dual Graphs of Polyhedral Decompositions for the Detection of
Adversarial Attacks}

\author{\IEEEauthorblockN{Huma Jamil}
\IEEEauthorblockA{\textit{Computer Science Department} \\
\textit{Colorado State University}\\
Fort Collins, CO, USA \\
huma97@colostate.edu}
\and
\IEEEauthorblockN{Yajing Liu}
\IEEEauthorblockA{\textit{Mathematics Department} \\
\textit{Colorado State University}\\
Fort Collins, CO, USA \\
yajing.liu@colostate.edu}
\and
\IEEEauthorblockN{Christina M. Cole$^*$}
\IEEEauthorblockA{\textit{Mathematics Department} \\
\textit{Colorado State University}\\
Fort Collins, CO, USA \\
christina.rigsby@colostate.edu}
\and 
\IEEEauthorblockN{Nathaniel Blanchard}
\IEEEauthorblockA{\textit{Computer Science Department} \\
\textit{Colorado State University}\\
Fort Collins, CO, USA \\
nblancha@colostate.edu}
\and
\IEEEauthorblockN{Emily J. King}
\IEEEauthorblockA{\textit{Mathematics Department} \\
\textit{Colorado State University}\\
Fort Collins, CO, USA \\
emily.king@colostate.edu}
\and
\IEEEauthorblockN{Michael Kirby}
\IEEEauthorblockA{\textit{Mathematics Department} \\
\textit{Colorado State University}\\
Fort Collins, CO, USA \\
michael.kirby@colostate.edu}
\and
\IEEEauthorblockN{Christopher Peterson}
\IEEEauthorblockA{\textit{Mathematics Department} \\
\textit{Colorado State University}\\
Fort Collins, CO, USA \\
christopher2.peterson@colostate.edu}

}
\maketitle
\begin{abstract}
Previous work has shown that a  neural network with the rectified linear unit (ReLU) activation function leads to a convex polyhedral decomposition of the input space. These decompositions can be represented by a dual graph with vertices corresponding to polyhedra and edges corresponding to polyhedra sharing a facet, which is a subgraph of a Hamming graph. This paper illustrates how one can utilize the dual graph to detect and analyze adversarial attacks in the context of digital images. 
When an image passes through a network containing ReLU nodes, the firing or non-firing at a node can be encoded as a bit ($1$ for ReLU activation, $0$ for ReLU non-activation). The sequence of all bit activations identifies the image with a bit vector, which identifies it with a polyhedron in the decomposition and, in turn, identifies it with a vertex in the dual graph.
We identify ReLU bits that are discriminators between non-adversarial and adversarial images and examine how well collections of these discriminators can ensemble vote to build an adversarial image detector. 
Specifically, we examine the similarities and differences of ReLU bit vectors for adversarial images, and their non-adversarial counterparts, using a pre-trained ResNet-50 architecture. While this paper focuses on adversarial digital images, ResNet-50 architecture, and the ReLU activation function, our methods extend to other network architectures, activation functions, and types of datasets.

\end{abstract}

\let\thefootnote\relax\footnotetext{978-1-6654-8045-1/22/\$31.00 \copyright{}2022 IEEE}

\section{Introduction}
A number of deep neural network models (e.g, Resnet-50 \cite{he2016deep}, DenseNet~\cite{huang2017densely}, Inception \cite{szegedy2016rethinking}, (to name a few) have demonstrated high accuracy on the classification of natural imagery.  As a result, neural networks are being increasingly exploited in workflows, e.g., in commerce, transportation, medicine, and threat detection, that need to be robust and resilient. Attempts to construct exemplars  deliberately designed to deceive these models are referred to as {\it adversarial attacks}.  Such attacks pose a significant potential threat to systems that leverage machine learning for pattern recognition and predictive analytics, such as self-driving cars \cite{abeysirigoonawardena2019generating}, or medical diagnostics \cite{ma2021understanding}.

There are a number of mechanisms to generate adversarial perturbations of a real image that can fool otherwise well-performing deep neural networks including convolutional neural networks (CNNs). \cite{szegedy_intriguing_2013,carlini_evaluating_2019,moosavi-dezfooli_deepfool_2016,DAmageNet_chen}.
In parallel, considerable research has been directed towards understanding the nature of adversarial images and how a network can be fooled \cite{pedraza_really_2022,ilyas_adversarial_2019,goodfellow_explaining_2015}, as well as building defenses against adversarial attacks \cite{meng_magnet_2017,athalye_obfuscated_2018,kotyan_adversarial_2022}.

\begin{figure}[t]
\centering
    \includegraphics[width = 0.99\linewidth]{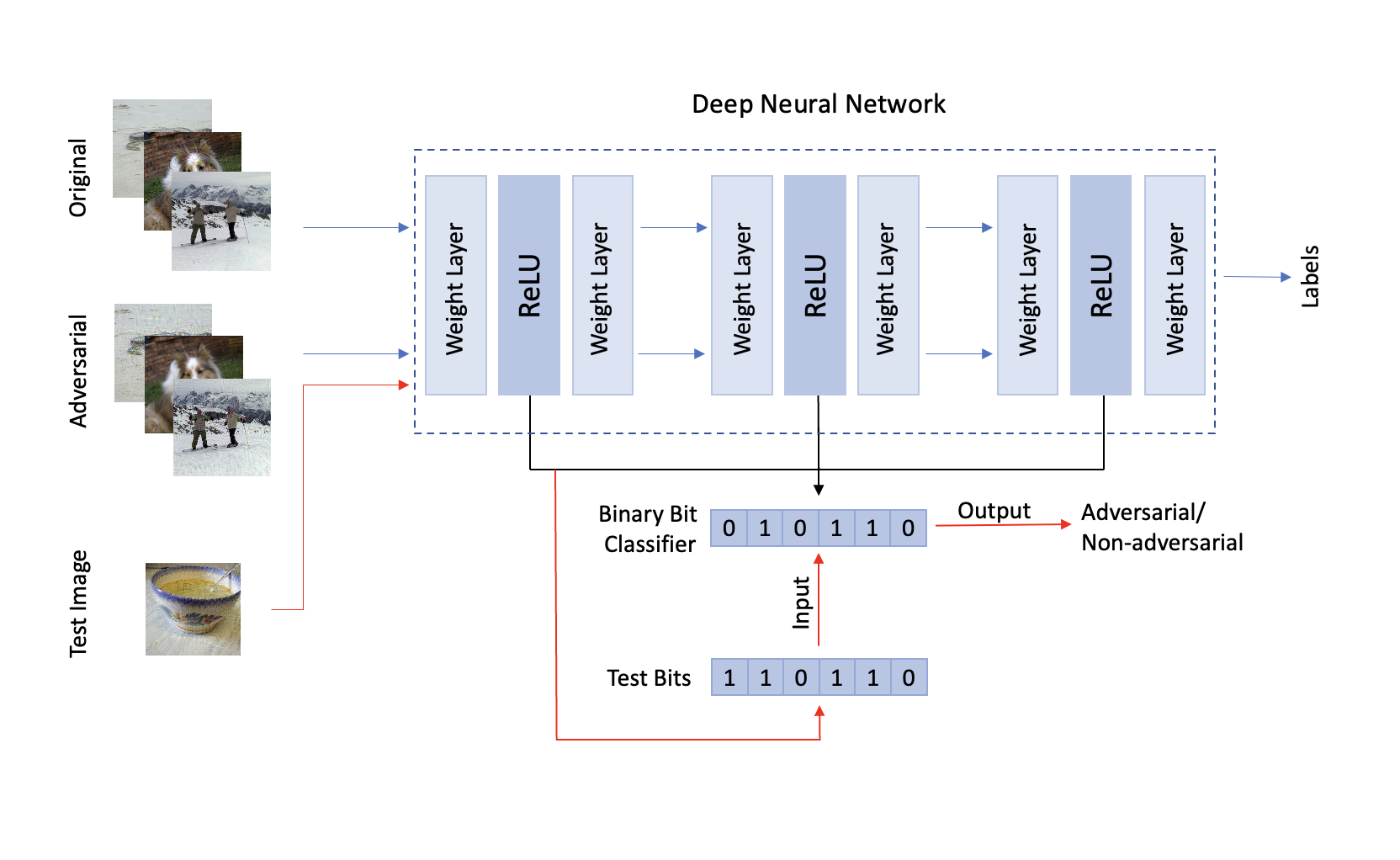}
    \caption{A high level overview of the construction of the binary bit classifier: the binary bits themselves are extracted from the ReLU activations from both adversarial and original image inputs. The bits are then used to construct the binary bit classifier, which can be used to classify a test image into either an adversarial or a non-adversarial category.}
    \label{teasorimage}
\end{figure}

To investigate the geometry of the data associated with adversarial attacks, one prominent area of research focuses on obtaining meaningful insights from the hidden layer representation of neural networks using metric learning \cite{mao_metric_2019} and examining the behaviors when projected to lower dimensional spaces through manifold learning \cite{jha_detecting_2018,stutz_disentangling_2019,lin_dual_2020}.
These papers show that in these lower dimensional manifolds, adversarial images  disentangle themselves and drift apart from original images or get closer to the targeted class in the case of targeted adversarial attack. This interesting behavior is then used to create various adversarial detection methods  \cite{crecchi_detecting_2019,pocos_examining_2022,wojcik_adversarial_2021,gorbett_utilizing_2022}.  

In this paper we present an alternative approach to characterizing and 
detecting adversarial images based on information in the dual graph
of a polyhedral decomposition.   

We propose the use of
bit vectors associated to training data and polyhedra partitioning the
input domain.  We provide evidence that these bit vectors
capture discriminatory information that distinguishes valid data from
that which has been adversarially perturbed.

This work has the following novel contributions: 
 \begin{itemize}
    \item a bit vector characterization based on the dual
    graph of a polyhedral decomposition to discriminate adversarial from natural images,
    \item  a classifier to identify discriminatory bits, and 
    \item an application of the approach to adversarial attacks on ResNet-50
    where adversarial images are identified with an accuracy rate around $90\%$. 

 \end{itemize}
Our work aims to relate adversarial images to bit vectors. In the future, we will investigate how the utilization of these bit vectors for adversarial image detection compares with current state-of-the-art methods. However, our current results focus on the exploration of bit vector properties and their link to adversarial images.

Other approaches in the literature connect polyhedral decompositions, or their associated bit vectors, to adversarial attacks but in different ways. For instance, Shamir et al., in~\cite{shamir_simple_2019} analyzed adversarial attacks with small Hamming distance, but with Hamming distance referring to the number of pixel values that were changed rather than the number of ReLU nodes that changed their activation parity. They found that one can change the classification determined by cells of a polyhedral decomposition or a trained neural network by changing a small number of pixels. Further, the polyhedral decompositions they consider are not generated by neural networks and thus have significantly fewer defining hyperplanes.

In~\cite{jin_efficient_2021}, the authors regularized training by Hamming distances between the bit vectors of perturbed data.  They showed that this approach yielded networks which were more robust to norm-bounded and perception-based adversarial attacks. In contrast, we are analyzing fixed neural networks not training new ones.

The paper is organized as follows: Section \ref{sec:review} discusses work related to the goals of this paper. Section \ref{sec:not-and-def} introduces  the ResNet architecture, the image datasets, and the definition of bit vectors. Section \ref{sec:link} first examines the similarities and differences of bit vectors for original images and their adversarial counterparts and then develops an algorithm to build the binary bit classifier.
Section \ref{sec:experiments} demonstrates that the developed binary bit classifier can be applied to discriminate adversarial images from non-adversarial ones with an accuracy rate around $90\%$. Section \ref{sec:future-work} describes the future work and Section \ref{sec:conclusion} concludes the paper.

\section{Geometric Framework}

\label{sec:review}

A trained feed forward neural network (FFNN) with the rectified linear unit (ReLU) as its activation function is a composition of a sequence of affine linear maps followed by component-wise application of ReLU, where ReLU leaves nonnegative values unchanged and maps negative values to zero.  
Suppose the neural network has $h$ ReLU nodes. We will associate to each point in the input space a vector in $\{0,1\}^h$ which we refer to as a {\it bit vector}. The entries of the bit vector record whether or not the corresponding ReLU node of the neural network fires for the given input point. The firing of the node can be interpreted as multiplying the number reaching the node by one while the non-firing of the node can be interpreted as multiplying the number reaching the node by zero.  Due to this interpretation, the entry in the bit vector corresponding to a given node is recorded as a one if the ReLU function fired and as a zero if the ReLU function did not fire. 

\begin{figure}[h]
\centering
     \includegraphics[width = 3.20in]
     {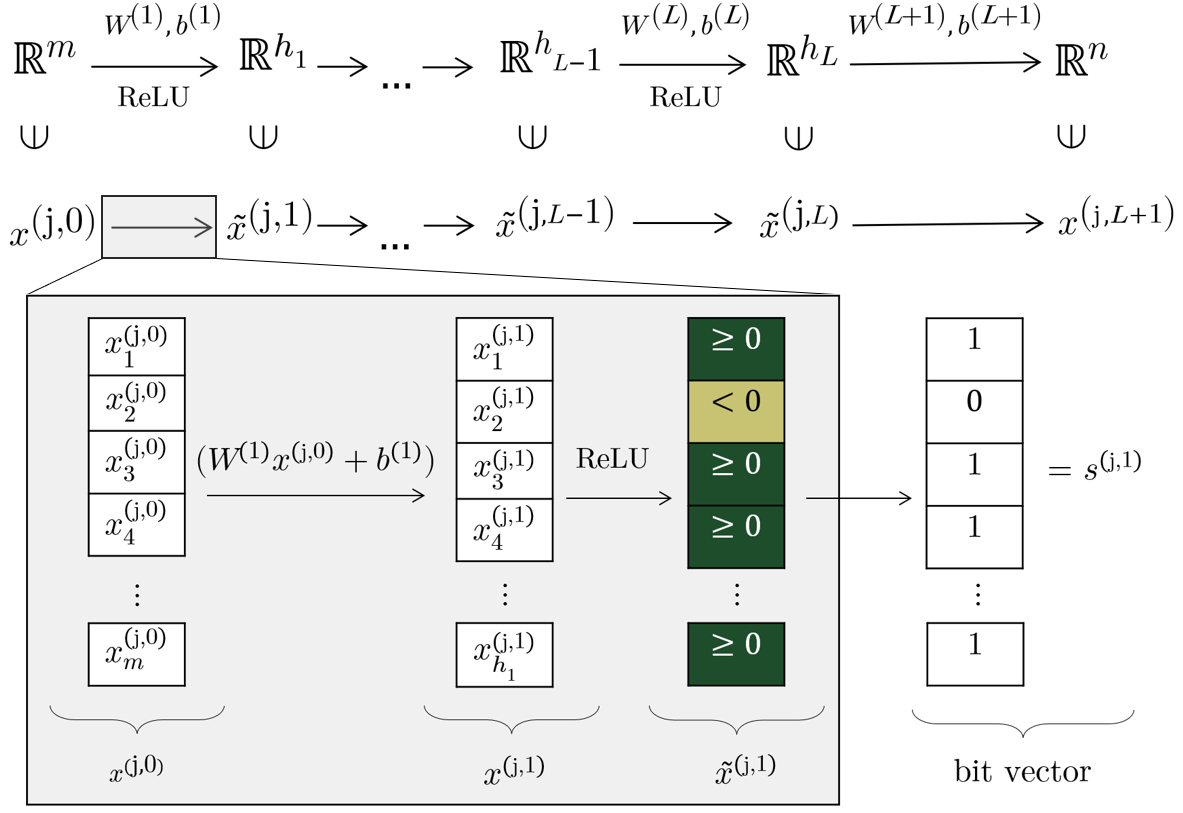}
    \caption{The process by which bit vectors are constructed from the layer-wise output of a ReLU FFNN using notation introduced in Section III(C).}
    \label{fig:bit_vec}
\end{figure}
    
A trained ReLU FFNN determines a convex polyhedral decomposition of its input space as each application of ReLU defines a hyperplane decision boundary in the input space. These decompositions can, in principle, be found using several algorithms \cite{xiang_reachable_2018,xu_traversing_2021,yang_reachability_2020}. To find the polyhedron containing a given point in input space, one must first find the bit vector corresponding to that point (see Fig. \ref{fig:bit_vec}). All points inside of a particular polyheron have the same associated bit vector, so we say that each polyhedron is labeled or uniquely identified by its bit vector. A priori, the number of distinct polyhedra would appear to be $2^h$. However, the actual number of polyhedra, which corresponds to the number of {\it realizable} bit vectors, is much smaller. Suppose we fix a neural network architecture and train the network on a fixed data set using a fixed training algorithm. If we vary the initial state of the network before training, the resulting neural networks after training will typically correspond to differing polyhedral decompositions with differing numbers of polyhedra. It is an open problem to understand the distribution of trained network outcomes corresponding to a given distribution of initializations. While the number of polyhedra that partition the input space varies under different initializations, the number can be shown to be much smaller than $2^h$ \cite{montufar_number_2014,pascanu_number_2013,serra_bounding_2018,hanin_deep_2019,arora_understanding_2018,hinz_framework_2019}. 
    
    As mentioned in the preceding paragraphs, each polyhedron in a given partition is uniquely identified by a bit vector. This effectively means that all points lying inside one of these polyhedra identically satisfy the inequalities defined by the layer-wise affine linear transforms (such as convolutions, multiplication of weight matrices, and addition of bias vectors) of the network and the ReLU activation pattern. More succinctly, the behavior of the neural network at points in a single polyhedron can be described with a single affine mapping from the network's input space to its output space \cite{sattelberg_locally_2020}. 
    
    We refer to two polyhedra as {\it neighbors} if they share a common facet. Hence, two polyhedra are neighbors if and only if their associated bit vectors differ in exactly one bit. As a consequence, the dual graph to the polyhedral decomposition is a (typically ``small'') subgraph of the Hamming graph corresponding to the possible ReLU activation patterns. Neighboring polyhedra have associated bit vectors that have Hamming distance one. The distance between two polyhedra can be measured as the length of a minimal walk in the dual graph of the decomposition between the corresponding vertices. Unfortunately, this way of measuring distances is impractical due to the size of the dual graph. The Hamming distance between the associated bit vectors of the polyhedra always serves as a lower bound for the minimal walk distance on the dual graph. Due to this property, we use the Hamming distance as a proxy for the minimal walk distance. Distance metrics derived from these polyhedral decompositions have been developed in~\cite{balestiero_spline_2018}.  Further, in~\cite{moser_tessellation_2022}, the topology of the Hamming subgraph induced from these polyhedral decompositions for networks trained on a simple (binary) classification task was explored.

\section{Model, Datasets, and Definitions}
\label{sec:not-and-def} 
In this section, we first introduce the ResNet architecture we use to derive the intermediate ReLU layer representations. Then we present the datasets we use in our analysis and experiments. Finally, we give the formal definition of bit vectors, the building blocks with which we construct our adversarial/non-adversarial classifier in Section~\ref{sec:link}.

\subsection {ResNet Architecture}

\begin{figure}[ht]
\centering
    \includegraphics[width=.99\linewidth]{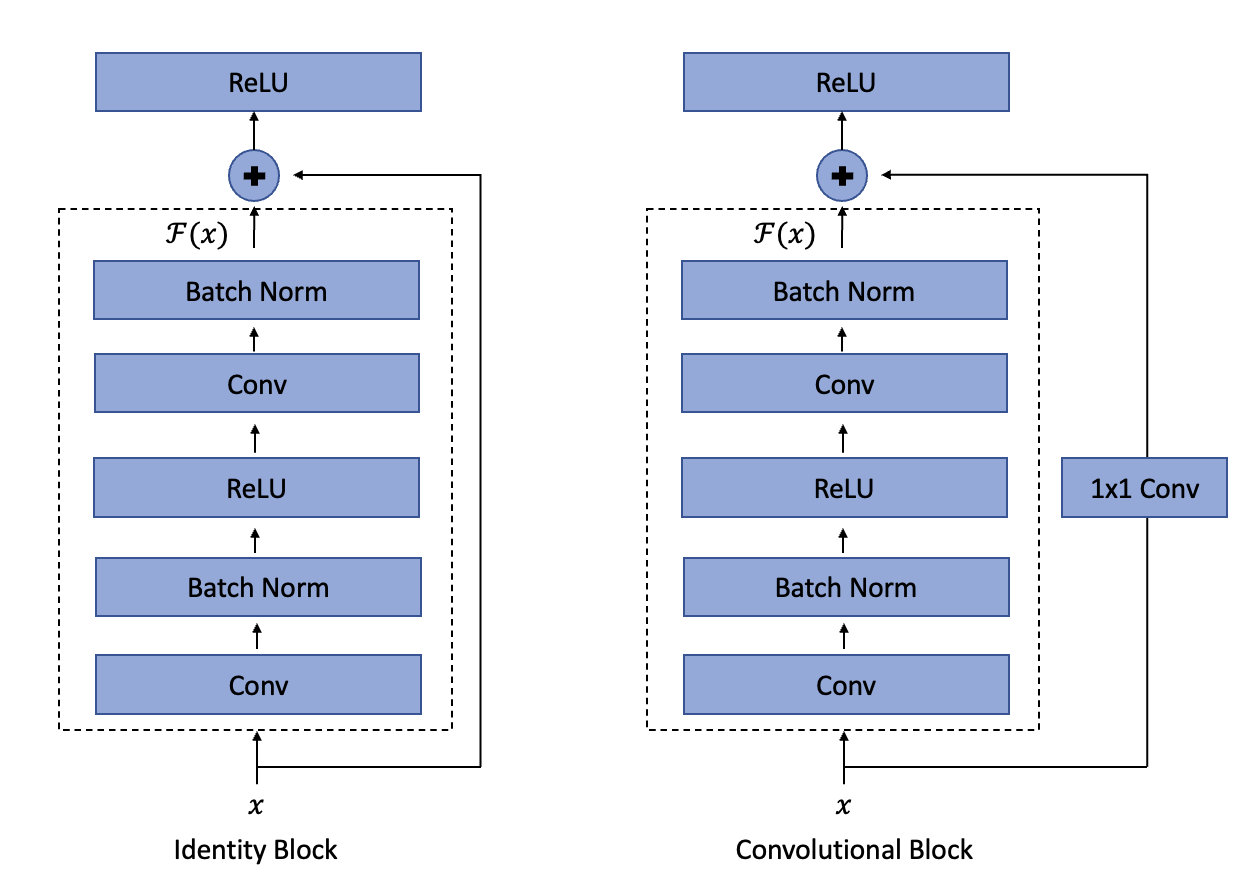}
    \caption{ResNet building blocks. }
    \label{Resnet}

\end{figure}

ResNet \cite{ResnetHe} is a popular neural network architecture. It is composed by stacking layers as blocks; each block includes convolutional layers, batch normalization, and ReLU activation functions. The mapping learned by each stack approximates the residual function $\mathcal{F}(x)$, which is then added to the input $x$. Further, the network is composed of shortcut connections that provide an alternative pathway for the gradients to flow during back-propagation, thus improving the degradation problem in the deep neural network. These shortcut connections can be made either by using an identity mapping or 1x1 convolution in case the sizes of $x$ and $\mathcal{F}(x)$ are different (Fig.~\ref{Resnet}).

Our experiments focus on one particular variant of ResNet: ResNet-50, a 50-layer convolutional neural network which is pre-trained on ImageNet~\cite{deng2009imagenet}. 
As discussed above, ResNet-50 uses ReLU as an activation function on the outputs from the convolutional layers; specifically, it has 17 ReLU layers in total.
In this paper, we use the bit vectors obtained through all 17 ReLU layers of ResNet-50 to investigate the similarities and differences between images from the ImageNet database and adversarial images created by two adversarial attack methods. Specifically, ResNet-50 takes images of input size 224x224x3, and the activation pattern of each ReLU layer is stored as a bit vector (see \eqref{eq:bitvec}). The total number of nodes for the first individual ReLU layer and each block for ResNet-50 is detailed in Table \ref{tab:numnodes_ReLU}. The novelty of this approach is that the computational complexity is greatly reduced by the use of bit vectors as compared to using embedding space features in 
$\mathbb{R}^m$ for some large dimension $m$.

Although we focused our analysis on ResNet-50, this technique could be applied to any neural networks with ReLU activation functions.

\begin{table}
\begin{center}
\begin{tabular}{ |c|c|c| }
\hline
Layers & Number of Outputs \\
\hline\hline
ReLU & 802816  \\  
\hline
Block1 (x3) & 802816 \\
\hline
Block2 (x4) & 401408 \\
\hline
Block3 (x6) & 200704 \\
\hline
Block4 (x3) & 100352 \\
\hline
\end{tabular}
\end{center}
\caption{Number of nodes in all 17 ReLU layers.}
\label{tab:numnodes_ReLU}
\end{table}
\subsection {Datasets}
\subsubsection{ImageNet}
ImageNet's validation dataset \cite{deng2009imagenet} contains 50000 images for 1000 classes. Each class has 50 images. Following typical practice, we centrally cropped and resized these images to a size of $224\times 224\times 3$ and then normalized them with mean and standard deviation obtained based on all images in the training dataset. 
\subsubsection{Adversarial images using FGSM}
We create an adversarial dataset, corresponding to the 50000 ImageNet images, using the Fast Gradient Sign method (FGSM)~\cite{goodfellow_explaining_2015}. 

To create an adversarial attack,  FGSM perturbs the original image in the direction determined by the gradients of the loss with respect to the input image thus maximizing the loss. It can be described using the following equation:
\begin{equation}
\label{eq: LGSM}
\hat{x} = x +\epsilon*\text{sign}(\nabla_x J(\theta, x, y)),
\end{equation}
where $x$ is the original input image, $\hat{x}$ is the adversarial image, $y$ is the original image label, $J$ is the loss, $\theta$ is the model parameter, and $\epsilon$ is the perturbation parameter.
To create this adversarial dataset we use the perturbation $\epsilon$ = 0.01. The adversarial images are cropped and normalized the same way as the ImageNet images. 
\subsubsection{DAmageNet}
DAmageNet is another  dataset containing 50000 universal adversarial samples generated from the ImageNet validation dataset. It was created by \cite{DAmageNet_chen} using an attack on attention: a technique which changes the attention heat map from the important area in the image to something irrelevant. This attack has shown to be very effective for multiple deep neural networks, even the networks with adversarial training can still have an error rate up to $70\%$ with DAmageNet attack \cite{DAmageNet_chen}.

\subsection{Bit Vectors}
The rectified linear unit (ReLU) function is a piecewise linear function defined as: 
\begin{equation*}
    \text{ReLU}(x) \coloneqq \max(x, 0) \ \text{for} \ x\in\mathbb{R},
\end{equation*}
where if 
$x=[x_1\ \ldots \, x_n]^{\top} \in\mathbb{R}^n$, then ReLU is applied on each component:
\textcolor{black}{
\begin{equation*}
    \text{ReLU}(x)=\begin{bmatrix}
    \text{ReLU}(x_1) \ 
    \ldots\ 
    \text{ReLU}(x_n) 
    \end{bmatrix}^{\top}.
\end{equation*}}
Due to its computational simplicity, representational sparsity, and linear behavior, a network that uses the ReLU activation function is easier to train and often achieves better performance \cite{nair2010rectified}. Therefore,
ReLU has become a default activation function in many types of neural networks, especially convolutional neural networks. 

Consider a  neural network which has the ReLU activation function occurring in some layers. For a given image $x^{(j,0)}$ from the neural network's input space $\mathbb{R}^m$, denote its output in the $i^{th}$ ReLU layer as $\Tilde{x}^{(j,i)}=[\Tilde{x}_1^{(j,i)} \ldots\ \Tilde{x}_{h_i}^{(j,i)}]^{\top} = \text{ReLU}(x^{(j,i)})\in\mathbb{R}^{h_i}$.
We define its bit vector in that layer as $s^{(j,i)} = [s_1^{(j,i)} \ldots\ s_{h_i}^{(j,i)}]^{\top}$ with 
\begin{equation}
\label{eq:bitvec}
    s_j^{(j,i)}\coloneqq \begin{cases}1 \ \text{if}\ \Tilde{x}_j^{(j,i)}>0\\
    0 \ \text{if}\ \Tilde{x}_j^{(j,i)} = 0,
    \end{cases}
\end{equation}
where $h_i$ is the number of nodes in the $i^{th}$ ReLU layer, $\sum_i h_i = h$. So, for an $h_i$-dimensional output $\Tilde{x}^{(j,i)}$ in the $i^{th}$ ReLU layer, $s^{(j,i)}$ gives a vector in $\{0,1\}^{h_i}$ encoding which nodes of the $i^{th}$ layer were activated by the ReLU function after taking $x^{(j,0)}$ as input. This is further clarified in Fig. \ref{fig:bit_vec} with a simple case of ReLU FFNN. These bit vectors represent the activation patterns that appear in the neural network as the input flows through it. Each region in the polyhedral decomposition of the input space follow the same path through the neural network, namely, the one described by the bit vector labeling that region.

\section{Link Between Bit Vectors and Images}
\label{sec:link}
In this section, we examine the link between bit vectors and image datasets.  We first investigate the similarities and differences between the original images from the ImageNet validation dataset and FGSM/DAmageNet adversarial images by analyzing their bit vectors in all 17 ReLU layers. Inspired by the differences, we then develop a binary classifier that can be used to detect adversarial images.
 \subsection{Similarities Between Original and Adversarial Images}
To the human eye, adversarially-altered images are very similar to their unaltered versions, but are often misclassified by neural networks. The bit vectors of original images and adversarial images do share some properties, as shown below.

\begin{figure}[ht]
\centering
    \includegraphics[width = 2.5in, height = 2in]{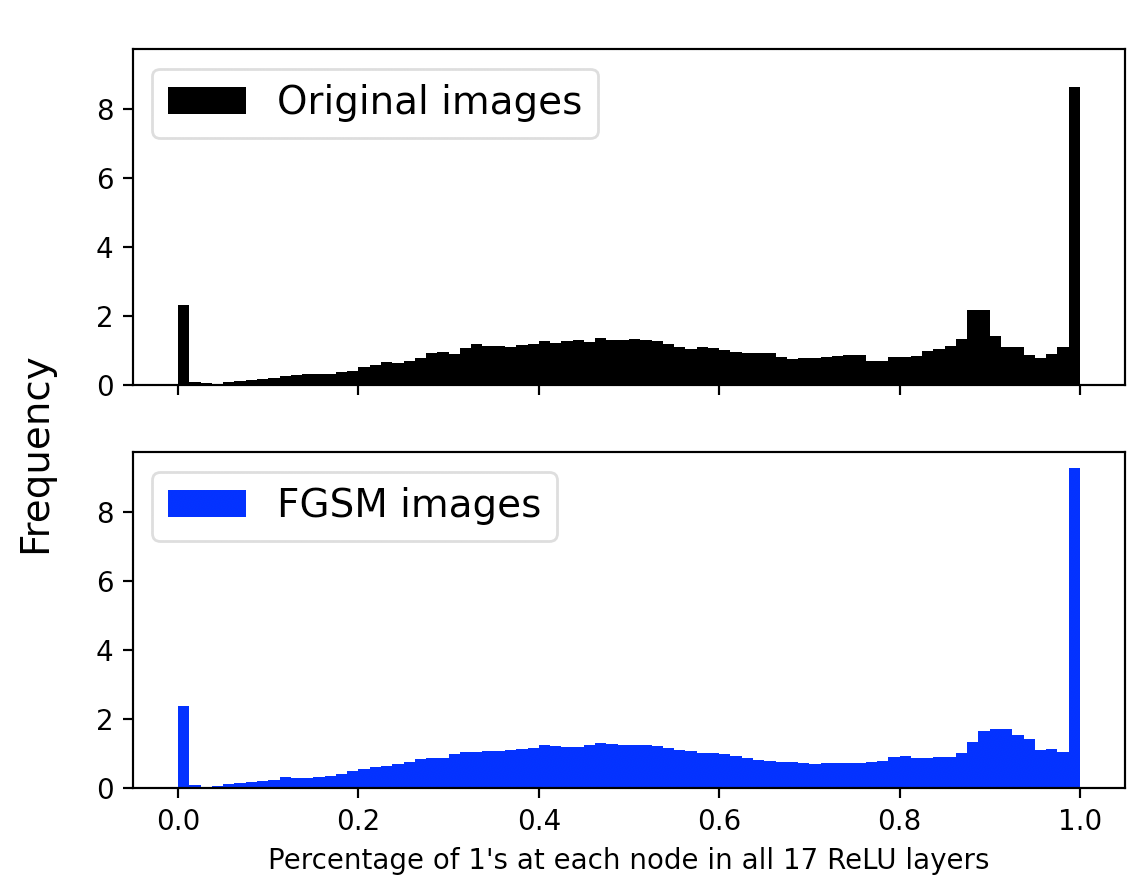}
    \caption{Distribution of percentage of active bits across all 17 ReLU layers for 50k original and FGSM images.}
    \label{distributionofbits}

\end{figure}

Fig.~\ref{distributionofbits} shows the distribution of the percentage of 1's at each node in all 17 ReLU layers for the 50k original and FGSM images, respectively. It can be observed that at many nodes all original images  have active bits (1's) while at some nodes all of them have inactive bits (0's), which is also true for FGSM images. This motivates us to ask whether the node indices where original images have 1's (or 0's) overlap with those where FGSM images also have 1's (or 0's). 
\begin{figure}[ht]
\centering
    \includegraphics[width = 2.6in, height = 2in]{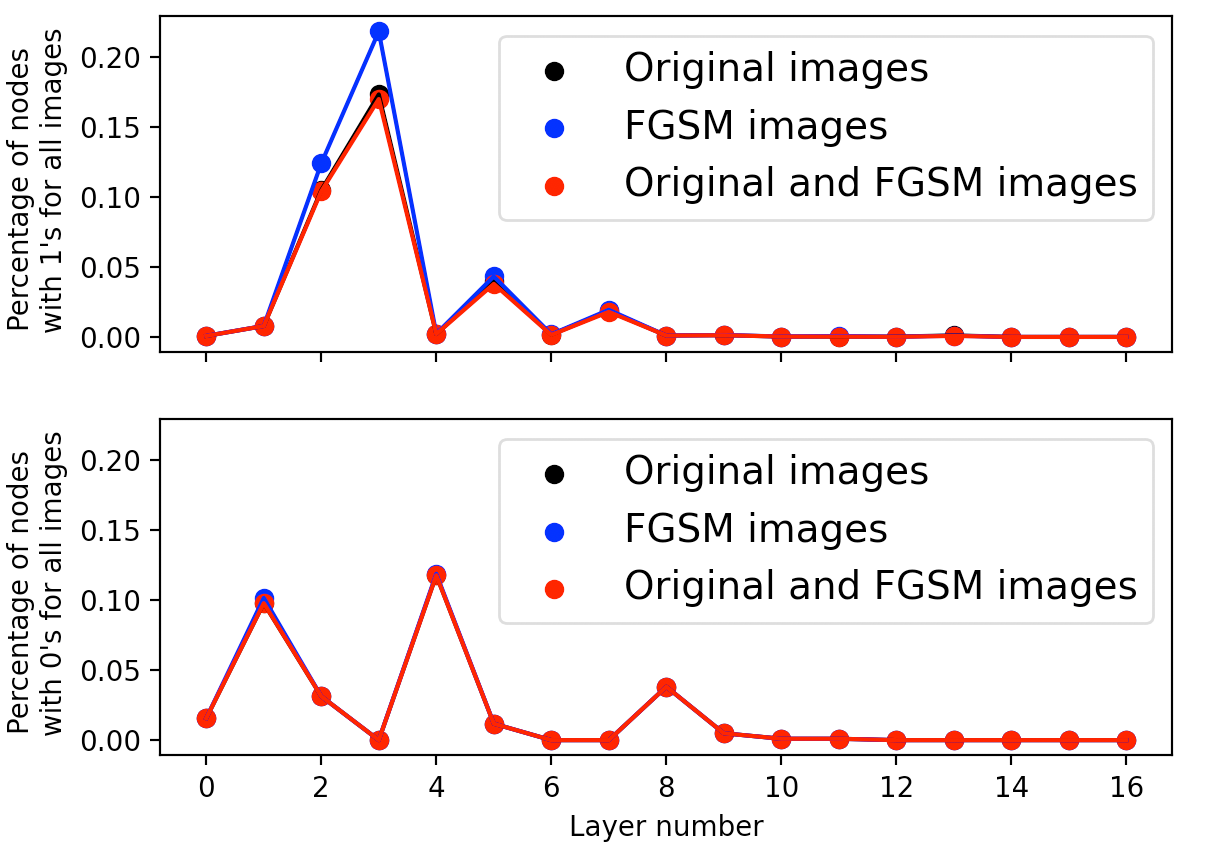}
    \caption{Percentage of nodes where all original/FGSM/original+FGSM images have common bits.}
    \label{common01_FGSM}
\end{figure}

\begin{figure}[ht]
\centering
    \includegraphics[width = 2.5in, height = 2in]{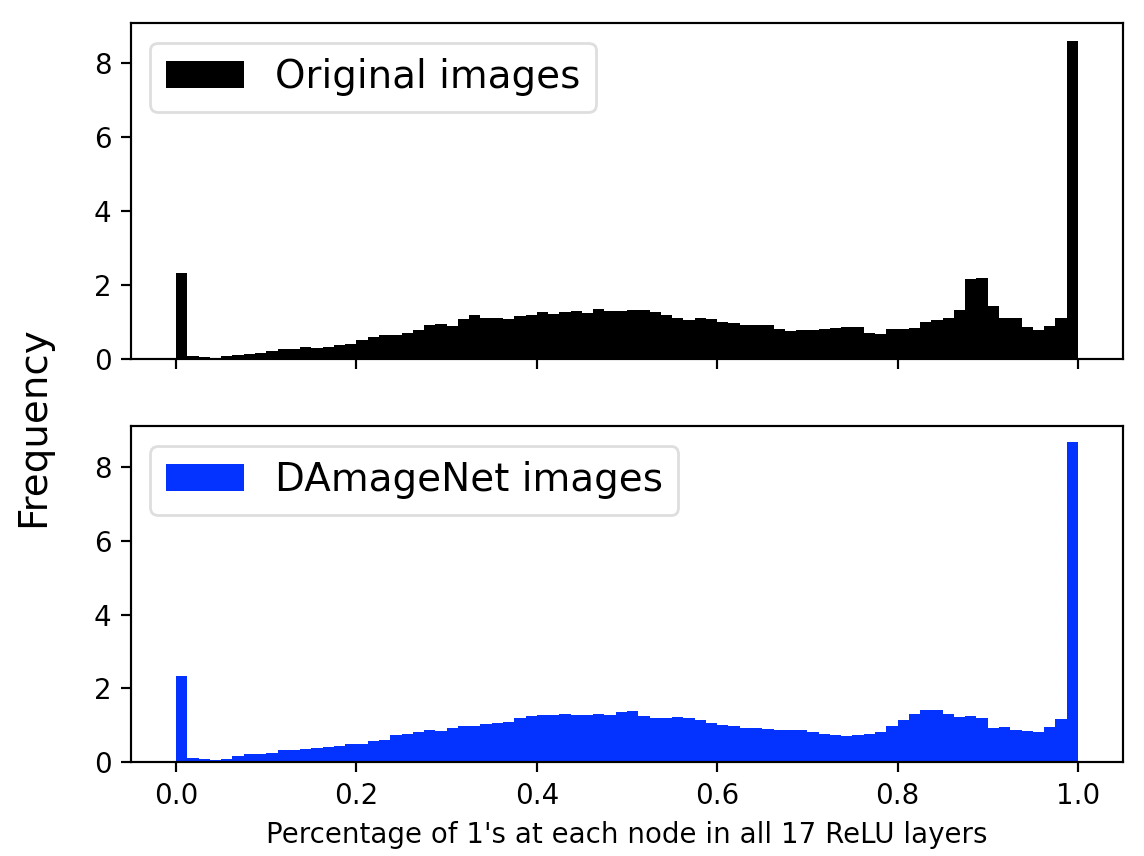}
    \caption{Distribution of percentage of active bits across all 17 ReLU layers for 50k original and DAmageNet images.}
    \label{distributionofbits_damage}

\end{figure}

\begin{figure}[ht]
\centering
    \includegraphics[width = 2.6in, height = 1.9in]{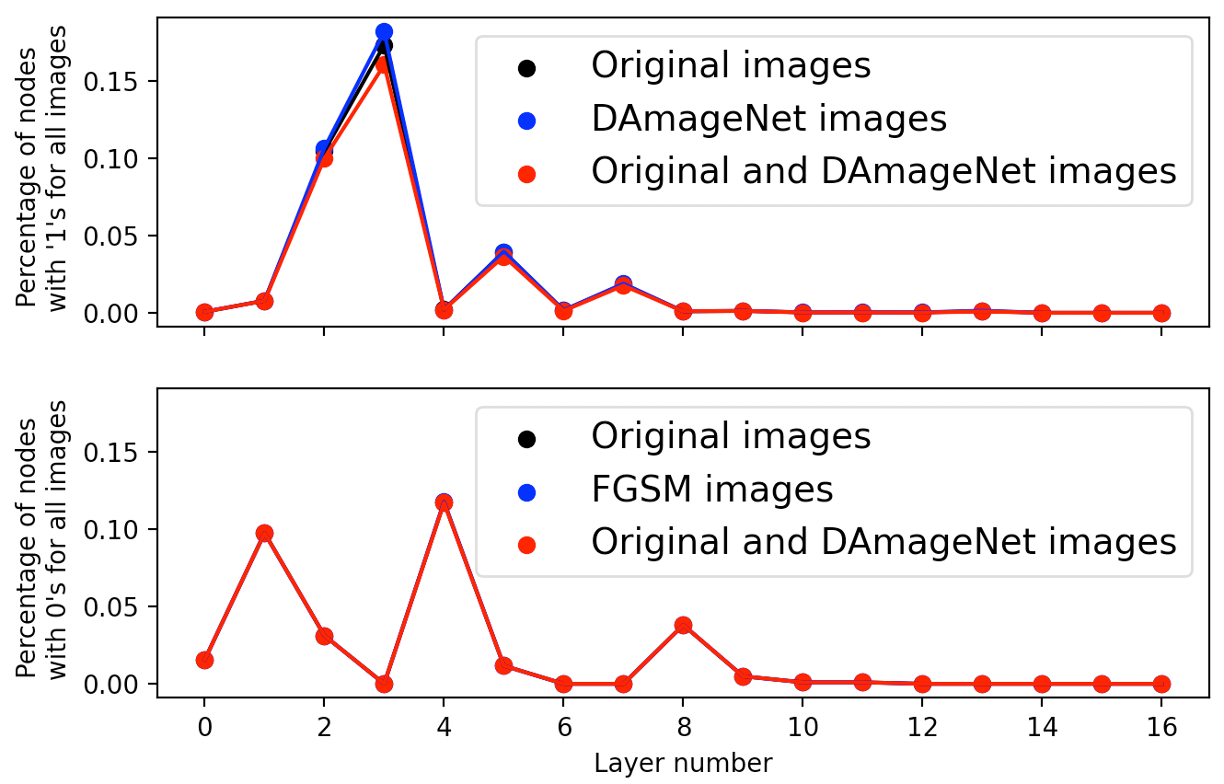}
    \caption{Percentage of nodes where all original/DAmageNet/original+DAmageNet images have common bits.}
    \label{common01_damagenet}
\end{figure}

The red line overlapping with the blue line for both plots in Fig.~\ref{common01_FGSM} shows that the answer to this question is `yes'. Further, Fig.~\ref{common01_FGSM} also shows that all original images have common 1's at more than $10\%$ of the nodes in ReLU layer 2 and more than $15\%$ of the nodes in ReLU layer 3, and they have common 0's at more than $10\%$ of the nodes in ReLU layers 1 and 4. No common bits exist exclusively for original images in the later ReLU layers. We observe the same trends between the original images and those altered with FGSM, but the number of common 1's shared among the altered images is greater than those shared among the original images. We make similar comparisons between the bit vectors of the original images and DAmageNet included in Fig.~\ref{distributionofbits_damage} and Fig.~\ref{common01_damagenet}, which reveal similar behavior as that seen between the original and FGSM images. 


\subsection{Differences Between Original and Adversarial Images}
In the intermediate layers of neural networks, the representations of the images change as the input propagates through each layer of the network. In the penultimate layer, the images are mapped to the embedding space that maps to the networks' final prediction (i.e., the softmax layer). We want to explore whether the bit vectors corresponding to the ReLU outputs across all the layers of the network, rather than just the penultimate layer, give meaningful information that can be used to differentiate adversarial images from the original ones.

\begin{figure}[ht]
\centering
     \includegraphics[width = 2.5in, height = 1.6in]{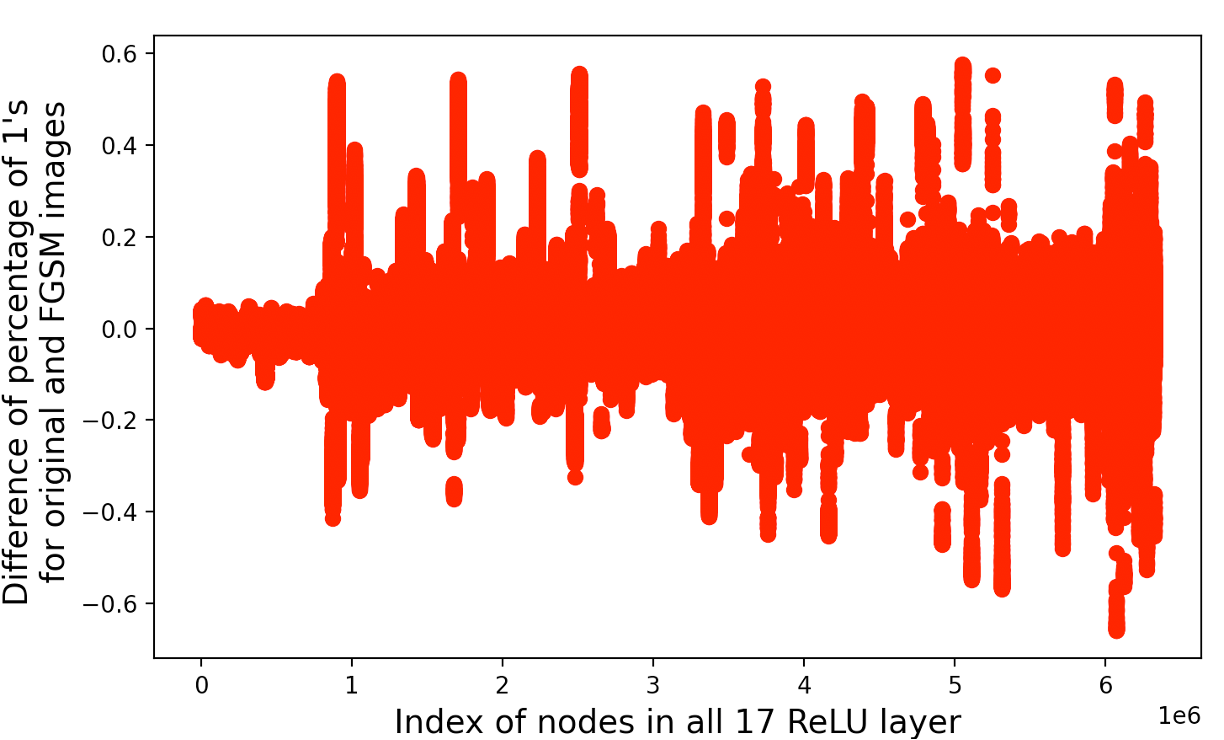}
    \caption{Difference of percentage of active nodes between original and FGSM images.}
    \label{plot2}
\end{figure}

One way to distinguish the adversarial images from the original ones would be to find discriminator nodes. Theoretically, these would be nodes where all original images and all adversarial ones are opposite, i.e., the output bits at those discriminators are $1$'s for all original images but $0$'s for all adversarial images (or, $0$'s for all original and $1$'s for all adversarial). As shown in Fig.~\ref{plot2}, such discriminators do not exist in practice (if they did, the difference would be $1$ or $-1$); however, Fig.~\ref{plot2} does show there are some nodes where the difference in the percentage of $1$'s across original ImageNet and FGSM images is around $\pm 0.6$. Such nodes may be distinct enough to approximate the theoretical power of perfect discriminator nodes. The differences $\pm 0.6$ in Fig.~\ref{plot2} imply that there may exist a subset of nodes where $80\%$ of original images have $1$'s (or $0$'s) and only  $20\%$ FGSM images have $1$'s (or $0$'s). This in turn motivates us to select special node indices where it is simultaneously true that a percentage higher than a certain threshold of the original images have $1$'s (or $0$'s) and more than the same threshold FGSM images have $0$'s (or $1$'s).  

\begin{figure}[ht]
\centering
    \includegraphics[width = 2.5in, height = 2in]{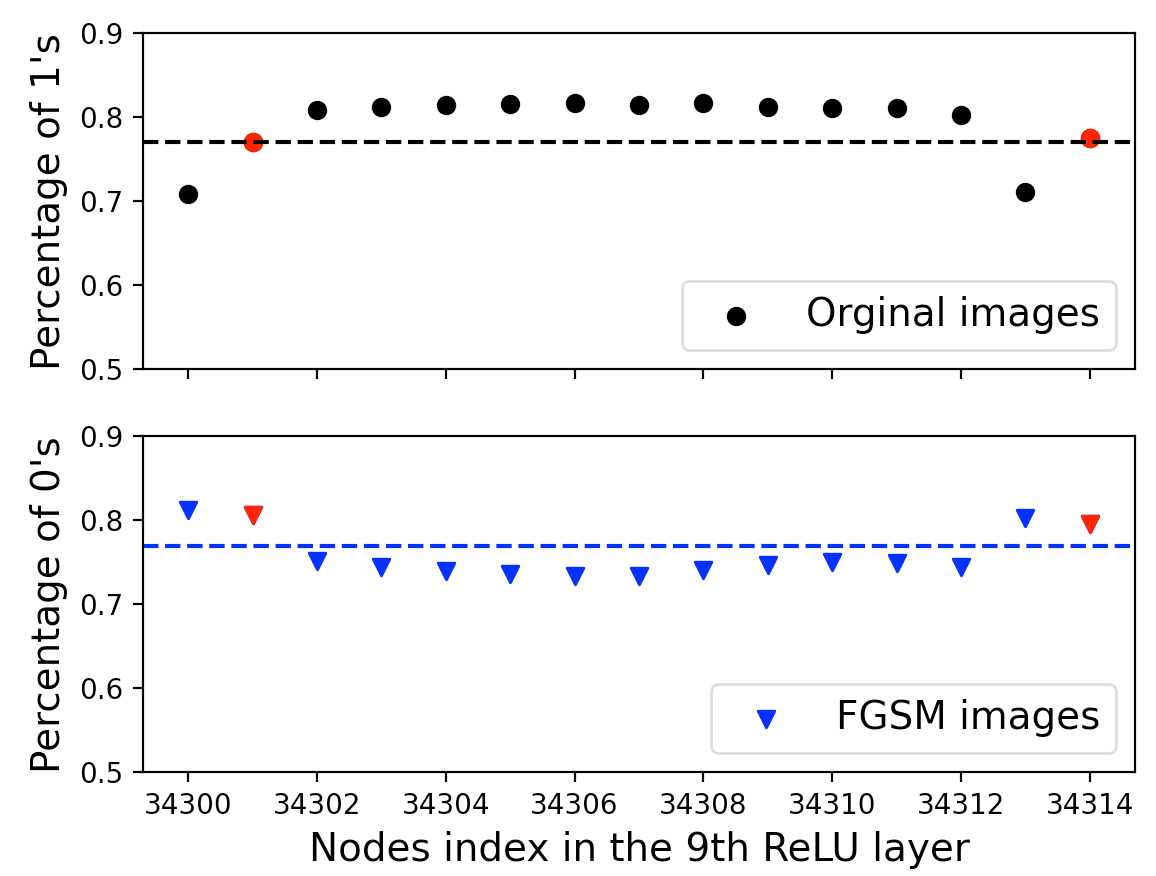}
    \caption{Selected node indices in the $9^{th}$ ReLU layer.}
    \label{filterednodes_layer9}
\end{figure}

\begin{figure}[ht]
\centering
    \includegraphics[width = 2.5in, height = 2in]{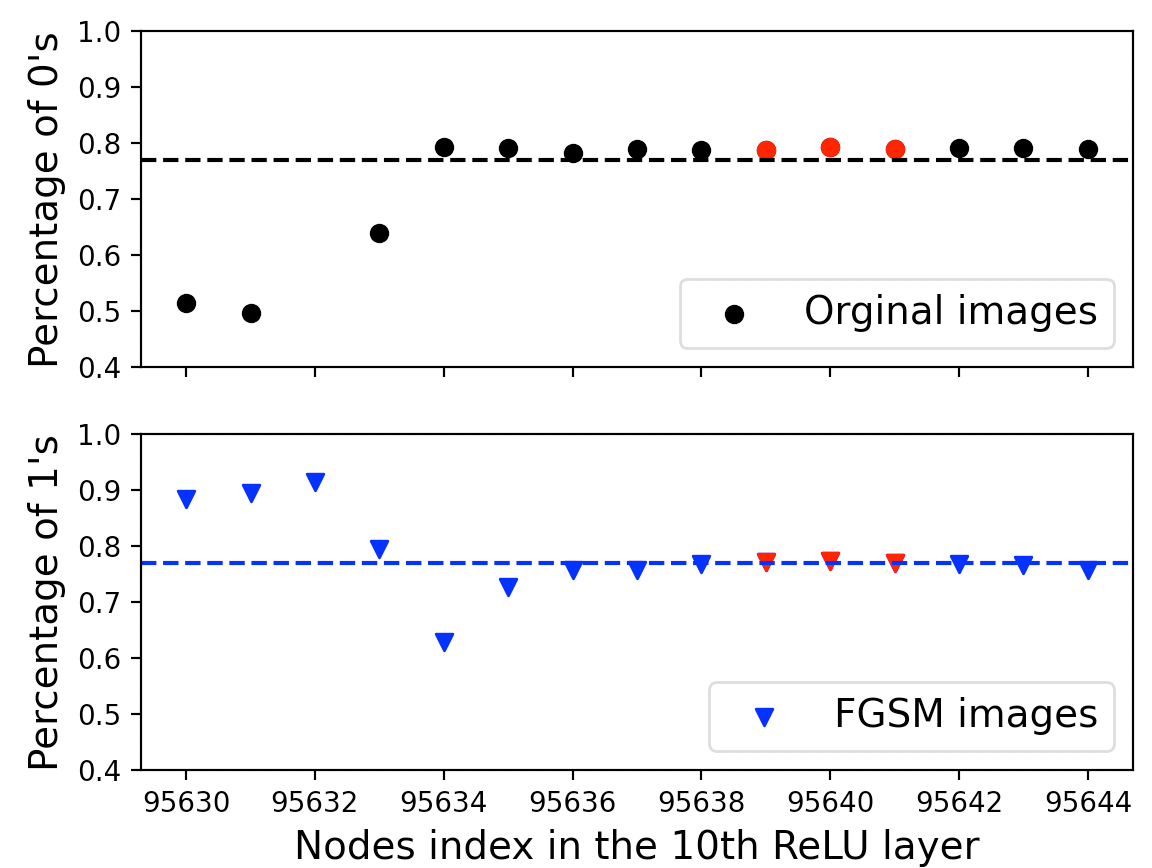}
    \caption{Selected  node indices in the $10^{th}$ ReLU layer.}
    \label{filterednode_layer10}
\end{figure}

For example, we set the threshold for the percentage of original images having $1$'s as $77\%$ to select a set of nodes and the same threshold for the percentage of adversarial images having $0$'s to select another set of nodes. We then take the intersection of these sets -- presuming it is non-empty -- to select the discriminator bits. Fig.~\ref{filterednodes_layer9} shows that from these two sets, there do indeed exist some discriminator bits, that is, two special nodes in the $9^{th}$ ReLU layer. Similarly, Fig.~\ref{filterednode_layer10} shows that there exist three special nodes where more than $77\%$ original images have $0$'s but more than $77\%$ adversarial images have $1$'s at those locations in the $10^{th}$ ReLU layer.

The selected indices inspire us to develop a binary classifier to differentiate adversarial images from original images by only using the selected special bit locations where the bits are different for the majority of original images and adversarial images and ``majority'' is defined by the threshold value.

\subsection{Binary Bit Classifier}
Let $\mathcal{O}$ and $\mathcal{A}$ be sets of $N$ original and adversarial images from given datasets. Using ResNet-50, we first extract the outputs of all intermediate 17 ReLU layers for each image. We then find its bit vectors in all 17 ReLU layers. Consider the $j^{th}$ original image, $x^{(j,0)}$ and the adversarially altered version of $x^{(j,0)}$, which we denote with $\hat{x}^{(j,0)}$. Define their bit vectors from the $i^{th}$ ReLU layer as $s^{(j,i)} = [s_{1}^{(j,i)} \ldots\ s_{h_i}^{(j,i)}]^{\top}$ and $\hat{s}^{(j,i)} = [\hat{s}_{1}^{(j,i)} \ldots\ \hat{s}_{h_i}^{(j,i)}]^{\top}$, respectively.

Of all nodes from ReLU layers, consider the $k^{th}$ one. Without loss of generality, say that this node is the $r^{th}$ node in the $i^{th}$ layer. We calculate $P_k$, the percentage of the original images whose bit vectors have a value of 1 in this location, and $\hat{P}_k$, the percentage of adversarial images whose bit vectors have a value of 1 in this location, which are defined 
\begin{align}
\label{eq:per_org}
    P_k &= \frac{1}{N}\sum_{j=1}^{N}s_r^{(j,i)}, \quad i = 1, \ldots, 17 \quad \textrm{and}\\
\label{eq:per_adv}
    \hat{P}_k &=  \frac{1}{N}\sum_{j=1}^{N}\hat{s}_r^{(j,i)}, \quad i = 1, \ldots, 17.
\end{align}

We then define a list of sets $C^{(i)} = [C^{(i)}_1, C^{(i)}_2, C^{(i)}_3, C^{(i)}_4], $ with
\begin{subequations}
\label{eq:C1}
\begin{align}
     C^{(i)}_{1}&=\{k: P_k \geq \lambda_{1}\} && (\lambda_{1}\in [0,1]),\\
    C^{(i)}_{2}&=\{k: 1-\hat{P}_k\geq \lambda_{2}\} && (\lambda_{2}\in [0,1]),\\
    C^{(i)}_{3}&=\{k: 1-P_k\geq \lambda_{3}\} && (\lambda_{3}\in [0,1]),\\
    C^{(i)}_{4}&=\{k: \hat{P}_k\geq \lambda_{4}\}&& (\lambda_{4}\in [0,1]).
\end{align}
\end{subequations}
Each of these four sets in the list $C^{(i)}$ denote the set of node indices at layer $i$ where the bits for at least $100\lambda_{1}$ percent original images are $1$'s, where the bits for at least $100\lambda_{2}$ percent adversarial images are $0$'s,   where the bits for at least $100\lambda_{3}$ percent original images are $0$'s, and  where the bits for at least $100\lambda_{4}$ percent adversarial images are $1$'s, respectively.\\
Now we define the list of sets $C = [C_A, C_B]$ with
\begin{equation}
    \label{eq:C_AC_B}
    \begin{aligned}
        C_A&= \Big(\bigcup_{i=1}^{17}C^{(i)}_{1} \Big) \cap \Big(\bigcup_{i=1}^{17}C^{(i)}_{2} \Big)\ \text{and}\\
        C_B&= \Big(\bigcup_{i=1}^{17}C^{(i)}_{3}\Big) \cap \Big(\bigcup_{i=1}^{17}C^{(i)}_{4}\Big),
    \end{aligned}      
\end{equation}
where $C_A$ denotes the set of node indices in all 17 ReLU layers where  the bits for at least $100\lambda_{1}$ percent original images are $1$'s, and at least $100\lambda_{2}$ percent adversarial images are $0$'s and $C_B$ denotes the set of node indices at all 17 ReLU layers where the bits for at least $100\lambda_{3}$ percent original images are $0$'s, and at least $100\lambda_{4}$ percent adversarial images are $1$'s. We denote the classifier bit vector as 
\begin{equation}
    \label{eq:classifiervector}
    B_C=[\underbrace{1\ \ldots\  1}_{|C_A|}\underbrace{\ 0\ \ldots\ 0}_{|C_B|}]^{\top},
\end{equation}
where $|C_A|$ and $|C_B|$ denote the number of node indices in set $C_A$ and $C_B$, respectively.

Now for any new image $x^{(j,0)}$, we first calculate its bit vectors for all 17 ReLU layers through the ResNet-50, and extract the values at the node indices in the $C$, the list of classifier index sets. Then we compare the bit values from the extracted node indices with $B_C$. If at least 50\% node indices have the same bits, it is classified as an original image,  otherwise it is classified as an adversarial image. The pseudo code for obtaining the binary classifier is summarized below in Algorithm~\ref{alg1}.

\begin{algorithm} 
\caption{Binary Bit Classifier Algorithm} 
\label{alg1}
\begin{algorithmic} 
    \REQUIRE $\mathcal{O}, \mathcal{A}$, $\lambda_{1},\lambda_{2}, \lambda_{3}, \lambda_{4}$, ResNet-50
    \ENSURE $C$ and $B_C$
    \STATE 1) Compute $s^{(j,i)}$ for each image $x^{(j,0)}\in\mathcal{O}$ and $\hat{s}^{(j,i)}$ for each image 
$\hat{x}^{(j,0)}\in\mathcal{A}$ in the $i^{th}$ ReLU layer using (\ref{eq:bitvec}).

\STATE 2) Compute and $P_k$ and $\hat{P}_k$ using (\ref{eq:per_org}) and (\ref{eq:per_adv}), respectively.

\STATE 3) Compute $C^{(i)}$ using (\ref{eq:C1}).

\STATE 4) Obtain $C=[C_A, C_B]$ using (\ref{eq:C_AC_B}) and then generate $B_C$ using (\ref{eq:classifiervector}).
\end{algorithmic}
\end{algorithm}

\section{Experiments on Adversarial Detection}
\label{sec:experiments}
In this section, we apply the binary classifier to detect adversarial images. 
To test the classifying strength of our classifier, we perform two experiments, one using the FGSM dataset and the other using DAmangeNet. ImageNet serves as the original image dataset for both experiments. For both experiments, we divide the data into train, validation and test split with (40k, 5k, 5k) images each from the ImageNet dataset and FGSM or DAmageNet datasets. We use the same value for the four thresholds $\lambda_{1},\, \lambda_{2}, \,\lambda_{3}$, and $\lambda_{4}$.

For the first experiment, using the training data, we first find the classifier bits with 12 threshold values uniformly selected from $[0.45, 0.77]$. We use $0.77$ as the upper limit of the threshold values, because no classifier bits could be found with value $\geq 0.78$. 

\begin{figure}[ht]
\centering
     \includegraphics[width = 2.5in, height = 2in]{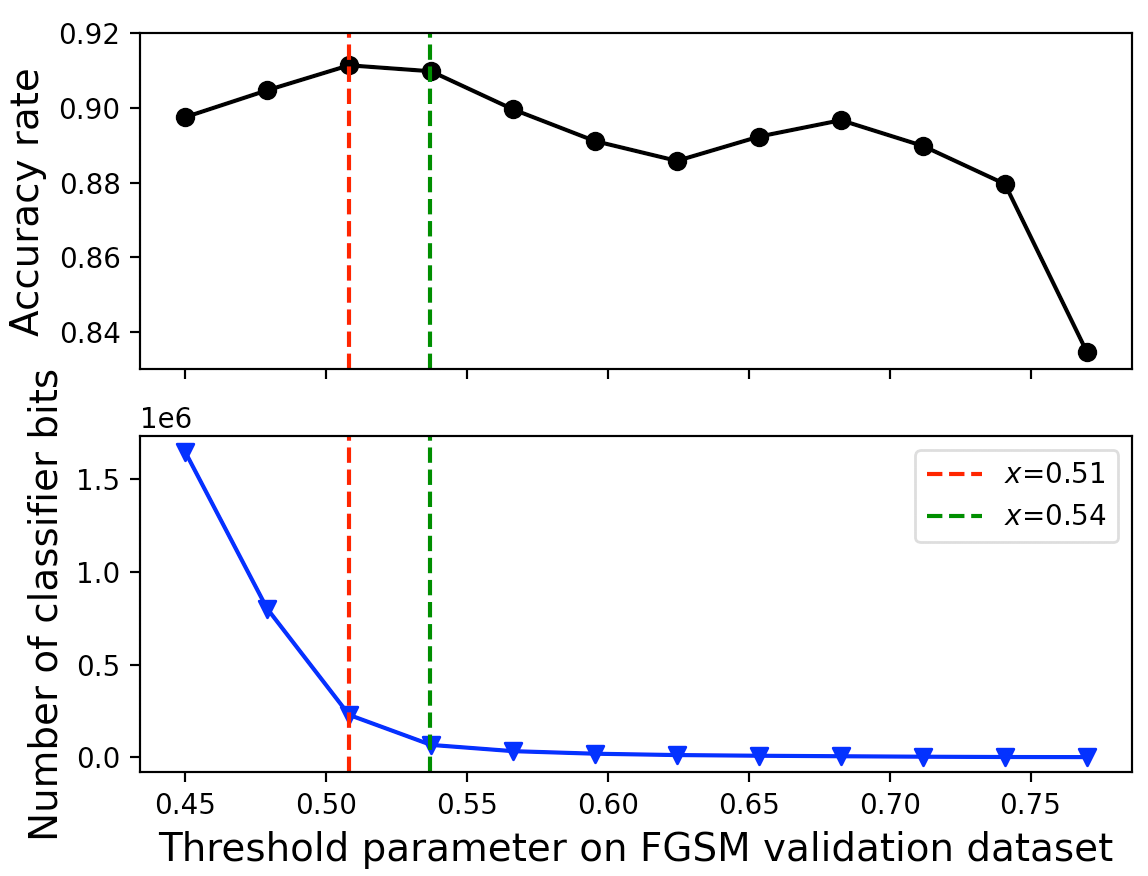}
    \caption{Accuracy plot for FGSM. }
    \label{hyperparameter_fgsm}
\end{figure}

After finding the classifier bits for different threshold values, we test the bits on the corresponding locations for the validation data to find the best threshold value based on the accuracy rate and the number of classifier bits. This threshold is then selected to check the accuracy of the classifier on the test images.  

Fig.~\ref{hyperparameter_fgsm} shows the performance of our binary classifier on the validation dataset for different threshold values and the corresponding number of classifier bits. The upper plot shows that we are able to achieve $83\%$ accuracy with the threshold value of $0.77$ which corresponds to only 5 classifier bits. The lower plot in Fig.~\ref{hyperparameter_fgsm} shows an initial sharp decrease in the number of classifier bit when we increase the threshold from $0.45$ to $0.51$, followed by a gradual decrease when we increase the threshold from $0.51$ to $0.77$. On the other hand, the upper plot in Fig.~\ref{hyperparameter_fgsm} shows that the accuracy of the classifier starts high and as we first increase the threshold, it initially increases before decreasing. The highest accuracy rate, $91.14\%$, is achieved at threshold $0.51$. This can be explained by the fact that weaker classifier bits can ensemble vote to augment the discriminatory capability of the strong classifier bits.

The accuracy for the threshold  0.54 is $90.98\%$, which is comparable to $91.14\%$. And the  true positive (TP) (where positive means non-adversarial) and true negative (TN) rates for these two threshold values are both comparable. The TP for $0.51$ and $0.54$ is 0.860 and 0.859, respectively, and the TN for them is 0.962 and 0.960, respectively. However, if we consider the number of classifier bits which influences the computational time, $0.54$ is a more suitable threshold to choose because it corresponds to far fewer classifier bits.

\begin{figure}[ht]
\centering
     \includegraphics[width = 2.5in, height = 1.8in]{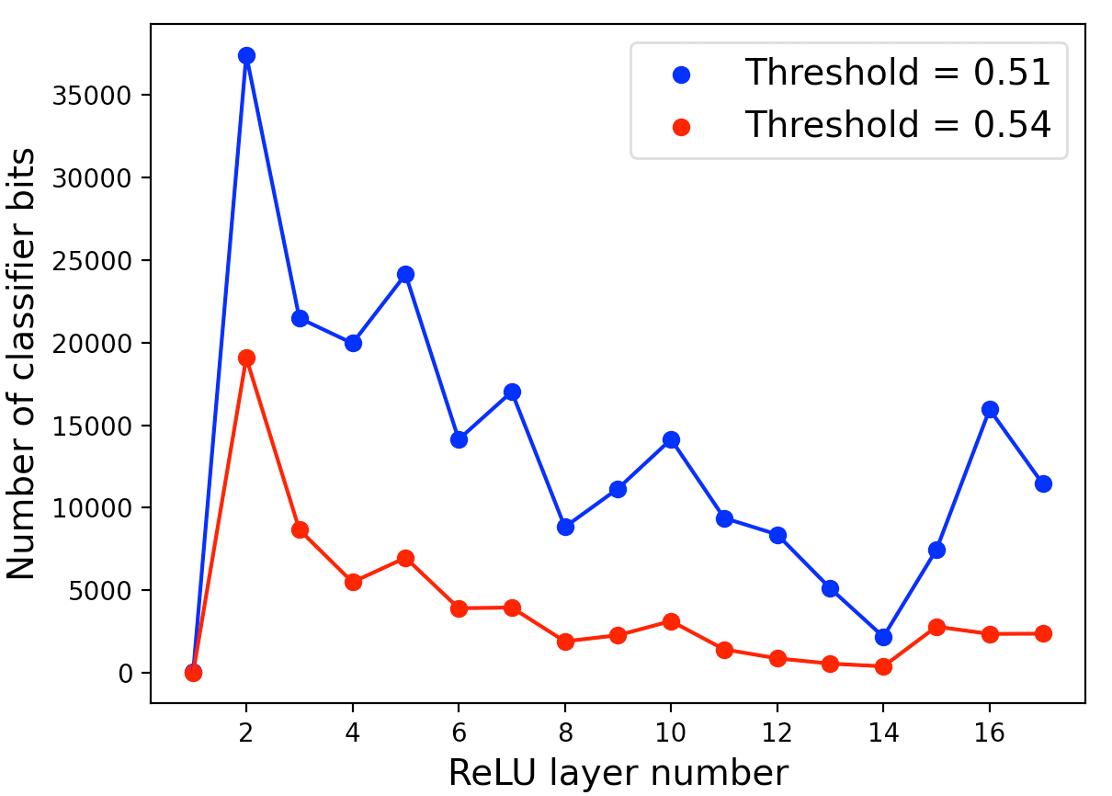}
    \caption{Classifier size comparison for threshold 0.51 and 0.54.}
    \label{classifierbits}
\end{figure}

Fig.~\ref{classifierbits} shows the number of classifier bits in all 17 ReLU layers for both thresholds. Layer 2 has the largest number of classifier bits among all layers for both parameters, but the threshold $0.51$ corresponds to a much bigger number of classifier bits in all layers than $0.54$. The accuracy rate for the test dataset is $89.84\%$ with the threshold $0.54$.

\begin{figure}[ht]
\centering
     \includegraphics[width = 2.5in, height = 1.6in]{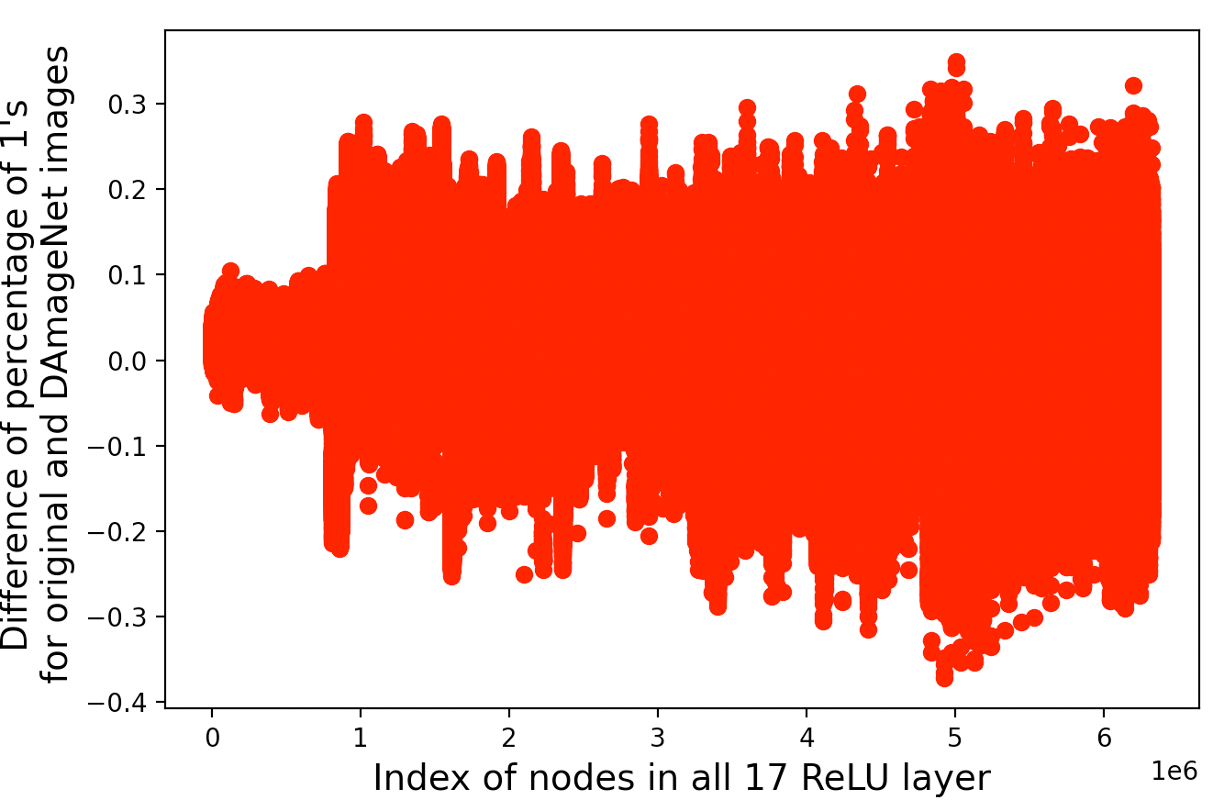}
    \caption{Difference of percentage of active nodes between original and DAmageNet images.}
    \label{subtraction1}
\end{figure}

The second experiment is performed with DAmageNet on similar lines with 11 threshold values uniformly selected from $[0.5,0.66]$. The upper limit of the threshold is smaller than for FGSM, which is justified by the nature of adversarial attacks used to create these datasets. Specifically, images in DAmageNet subjectively appear more perturbed than FGSM images. Fig.~\ref{subtraction1} (cf.\ Fig.~\ref{plot2}) shows that the difference of the percentage of $1$'s is condensed between $-0.4$ and $0.4$; so, in order to find the classifier bits where the majority of the adversarial images behave exactly opposite to majority of the original images, we need to lower the threshold.

\begin{figure}[ht]
\centering
     \includegraphics[width = 2.5in, height = 2in]{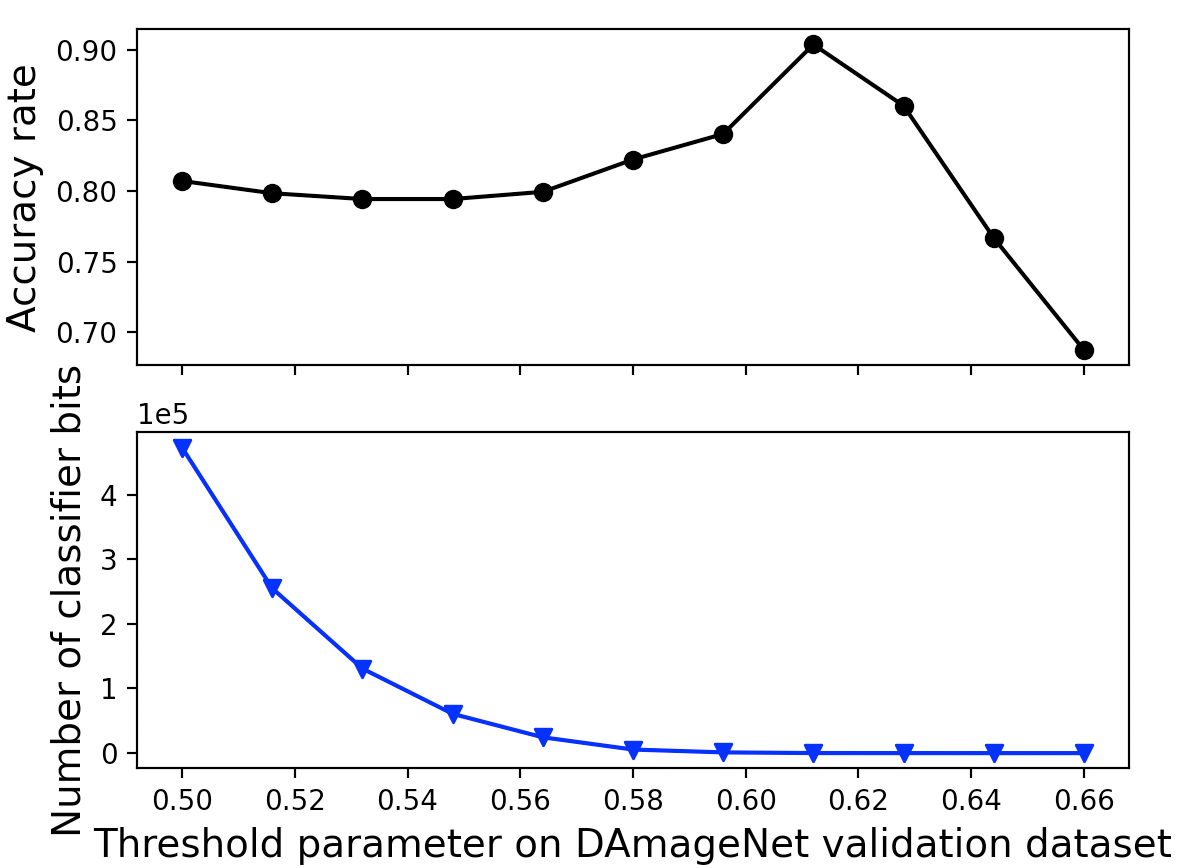}
    \caption{Accuracy plot for DAmageNet.}
    \label{accuracydamagenet}
\end{figure}

For DAmageNet, we observe (see Fig.~\ref{accuracydamagenet}) the peak accuracy $90.38\%$ at the threshold of $0.61$ and then the accuracy drops significantly if we decrease the threshold value. Also, for all thresholds, the number of classifier bits are one order of magnitude less than the FGSM. The accuracy rate for the test dataset is $90.81\%$ with the threshold $0.61$.

\section{Future Work}\label{sec:future-work}
Future work will consider the following questions:
\begin{itemize}
    \item How should the ensemble voting of adversarial bit detectors be combined for stronger results?
    \item How well does the approach work on other adversarial data sets?
    \item Does our method specifically detect adversarial images or any out of distribution image?
    \item How does the utilization of these bit vectors for adversarial image detection compare with existing methods (for example, \cite{QinDetecting2019})?
    \item Can the geometry and topology of the Hamming subgraph associated to the bit vectors be further exploited?
    \item How robust is the adversarial bit vector approach on other network architectures?
\end{itemize}
The last of these subtly implies the generalizability of our work. Indeed, the context on which we have focused is neural networks with ReLU activation functions, but we note that other activation functions can be used. Given an input image, ReLU provides a natural discretization of the layer-wise outputs into bit vectors and is computationally inexpensive to implement. These bit vectors could, however, consist of integers beyond only 0 and 1 to bin the layer-wise outputs by magnitude. For example, if the sigmoid activation function is used, entries of layer-wise outputs before activation could be mapped to 0 if their values are less than or equal to -4, to 2 if their values are greater than or equal to 4, and to 1 for values between -4 and 4. Therefore, any activation function can be used if a complementary discretization map/binning method for the layer-wise output is defined. This map would be used to create vectors that describe particular paths through the neural network, vectors that partition the input space as the bit vectors defined herein do. We plan to use this insight to experiment with different architectures and activation functions in the future.

\section{Conclusion}\label{sec:conclusion}
The main contribution of this work is the link of ReLU activation patterns encoded as $\{0,1\}$ vectors to various kinds of adversarial attacks. We compared the ReLU activation patterns of adversarial images to the ReLU activation patterns of non-adversarial images to find the bit locations that have discriminatory capability of identifying adversarial attacks. These bit locations were then used to ensemble vote to construct a binary bit classifier. With this simple technique, we were able to distinguish between adversarial and non-adversarial images with over 90\% accuracy. This indicates the feasibility of utilizing dual graphs to analyze and detect adversarial attacks on images via the bit vectors. We believe our detection approach can be extended to other networks and adversarial attacks.

\section*{Acknowledgements}
This work is partially supported by the United States Air Force under Contract No. FA865020C1121 and the DARPA Geometries of Learning Program under Award No. HR00112290074.



\end{document}